\documentclass[journal]{IEEEtran}
\usepackage{amsmath,amsfonts}
\usepackage{algorithmic}
\usepackage{algorithm}
\usepackage{array}
\usepackage[caption=false,font=normalsize,labelfont=sf,textfont=sf]{subfig}
\usepackage{textcomp}
\usepackage{stfloats}
\usepackage{url}
\usepackage{verbatim}
\usepackage{graphicx}
\begin{document}

\title{Multi-Image Super Resolution Framework for Detection and Analysis of Plant Roots}

\author{Shubham Agarwal,~\IEEEmembership{Student Member,~IEEE}, Ofek Nourian, Michael Sidorov, Sharon Chemweno, Ofer Hadar, ~\IEEEmembership{Senior Member,~IEEE}, Naftali Lazarovitch, and Jhonathan E. Ephrath
\thanks{Manuscript received June 30, 2025; Revised October 20, 2025 and November 16, 2025; \textit{Corresponding Author: Shubham Agarwal.}}
\thanks{Shubham Agarwal, Ofek Nourian, Michael Sidorov and Ofer Hadar are with the School of Electrical and Computer
Engineering, Ben Gurion University of the Negev, Beersheba 84105, Israel
(e-mail: agarwals@post.bgu.ac.il, ofekno@post.bgu.ac.il, sidorov@post.bgu.ac.il, hadar@bgu.ac.il)}
\thanks{ Sharon Chemweno, Naftali Lazarovitch, and Jhonathan E. Ephrath are with
the French Associates Institute for Agriculture and Biotechnology of Drylands,
The Jacob Blaustein Institutes for Desert Research, Ben-Gurion University
of the Negev, Sede Boqer 84105, Israel (e-mail: chemweno@post.bgu.ac.il;
lazarovi@bgu.ac.il; yoni@bgu.ac.il).} }

\markboth{IEEE Transactions on AgriFood Electronics,~Vol.~X, No.~Y, Month~2025}{Agarwal \MakeLowercase{\textit{et al.}}: Multi-Image Super Resolution Framework for Detection and Analysis of Plant Roots}

\IEEEpubid{0000--0000/00\$00.00~\copyright~2025 IEEE}

\maketitle

\begin{abstract}
Understanding plant root systems is critical for advancing research in soil–plant interactions, nutrient uptake, and overall plant health. However, accurate imaging of roots in subterranean environments remains a persistent challenge due to adverse conditions such as occlusion, varying soil moisture, and inherently low contrast, which limit the effectiveness of conventional vision-based approaches. In this work, we propose a novel underground imaging system that captures multiple overlapping views of plant roots and integrates a deep learning-based Multi-Image Super-Resolution (MISR) framework designed to enhance root visibility and detail. To train and evaluate our approach, we construct a synthetic dataset that simulates realistic underground imaging scenarios, incorporating key environmental factors that affect image quality. Our proposed MISR algorithm leverages spatial redundancy across views to reconstruct high-resolution images with improved structural fidelity and visual clarity. Quantitative evaluations show that our approach outperforms the state-of-the-art (SOTA) super resolution baselines, achieving 2.3\% reduction in BRISQUE indicating improved image quality with same CLIP-IQA, thereby enabling enhanced phenotypic analysis of root systems. This in turn facilitates accurate estimation of critical root traits, including root hair count and root hair density.  The proposed framework presents a promising direction for robust automatic underground plant root imaging and trait quantification for agricultural and ecological research.
\end{abstract}

\begin{IEEEkeywords}
Multi-Image Super Resolution, Plant Root Analysis, Object Detection, Image Enhancement, Underground Camera System, Simulated Plant Root Dataset
\end{IEEEkeywords}

\section{Introduction}

With limited resources, depleting soil quality and increasing demand for food, artificial intelligence (AI) in agriculture is emerging as an important area of research. There have been several deep learning based algorithms that analyze plant characteristics like leaf diseases identification, insect attacks, fruit quality etc. Aerial imagery using drones is also used to monitor forest and agricultural land areas \cite{dron1, dron2}. These methods greatly advance our knowledge about plant growth to get better yield. One of the areas that is comparatively less explored is the analysis of underground plant roots. Roots have an essential role in the overall well-being of the plant \cite{rootb1, rootb2, rootb3, rootb4}. Apart from providing anchorage, they absorb water and nutrients from the soil. Roots also have hairs that help in more efficient absorption. Analysis of plant roots is difficult as they are underground and imaging systems face numerous difficulties like moist soil, occlusion and low contrast \cite{diff1, diff2}. 

\IEEEpubidadjcol

In this work, we investigate the application of multi-image super-resolution (MISR) technique to enhance the quality of underground RGB images for plant root analysis. To facilitate this, we use a hardware system, termed Multi-Image RootCam, specifically designed to capture high-resolution, overlapping RGB images of plant roots in subterranean environments. The Multi-Image RootCam setup enables the acquisition of multiple spatially overlapping views of the root structure, which serve as inputs for our MISR pipeline.

In addition to real image acquisition, we construct a synthetic dataset that simulates the overlapping underground root images produced by Multi-Image RootCam. This simulated dataset is employed to train our proposed MISR model, enabling controlled experimentation and scalable model development. Our MISR architecture builds upon the state-of-the-art (SOTA) Deep Residual Channel Attention Transformer (DRCT) model, originally developed for single-image super-resolution tasks. We extend and adapt this architecture to process multiple overlapping image inputs, leveraging the complementary information present across views to enhance resolution more effectively than single-image approaches. Through extensive experiments, we demonstrate that our model significantly outperforms existing super-resolution techniques in terms of both quantitative metrics and visual fidelity. 

The main contributions of this work are summarized as follows:
\begin{itemize}
\item We design a novel RootCam imaging system capable of acquiring overlapping underground RGB views of plant roots for multi-image super-resolution.
\item We introduce a synthetic overlapping root dataset that simulates underground imaging conditions for robust model training.
\item We propose the MI-DRCT framework, which integrates a multi-image fusion and alignment module into the DRCT backbone to exploit subpixel shifts across overlapping views.
\item The proposed approach demonstrates significant quantitative and perceptual gains, improving root structure clarity and downstream segmentation accuracy.
\end{itemize}

Our framework achieves remarkable results in enhancing the root image quality surpassing the closest method by 2.3\% on the BRISQUE \cite{BRISQUE} score with same CLIP-IQA \cite{clipiqa} score.
 The results indicate the superiority of our method when applied to underground root image analysis. 

The enhanced image quality provided by our system enables better quantification of critical root phenotypic traits, such as root hair count and surface area. By integrating the Multi-Image RootCam hardware with our improved MISR pipeline, we achieve detailed visualization of subterranean plant roots, which holds significant potential for advancing root phenotyping research.

\section{Related Works}

\subsection{Super Resolution}

Classical multi-image super-resolution (MISR) approaches date back to the 1970s \cite{peleg1987improving}, where multiple low-resolution (LR) images were fused to correct aliasing artifacts. While effective for small-scale improvements, these early methods relied heavily on handcrafted priors and lacked robustness to noise and motion inconsistencies.

With the advent of transformers \cite{vaswani2017attention}, their success in capturing global dependencies led to rapid adoption in vision tasks \cite{dosovitskiy2020image, liu2021swin}. Consequently, numerous works explored transformer-based architectures for super-resolution (SR) \cite{trans1, trans2, trans3, trans4, trans5, trans6}. However, the transition from CNN-based local modeling to transformer-based global reasoning remains nontrivial, often leading to high computational costs and unstable convergence when training on limited data.

Yang et al. \cite{yang2020learning} proposed a texture transformer to mitigate texture degradation in SR by encoding texture features as transformer keys, values, and queries. While this approach effectively recovers fine details, it depends heavily on reference images and suffers when such references are unavailable or inconsistent. Similarly, Chen et al. \cite{chen2021pre} introduced the Image Processing Transformer (IPT), a large-scale pre-trained model for low-level tasks. Although IPT demonstrated impressive results, its practicality is limited by its enormous size (115M parameters) and reliance on large multi-task datasets, restricting deployment in resource-constrained environments.

For video SR, Cao et al. \cite{caoetal} extended transformer-based reasoning to spatiotemporal data via VSR Transformer. However, their reliance on CNN-derived features limited the model’s ability to capture long-term temporal dependencies, and the patch-wise attention mechanism caused local context fragmentation. Liang et al. \cite{liang2021swinir} adapted Swin Transformer \cite{liu2021swin} for SR by combining CNN-based shallow features with transformer-based deep representations. Although this hybrid structure improved local feature translation, it still struggled with computational efficiency due to hierarchical window shifting.

The Uformer model \cite{wang2022uformer} adopted a U-shaped transformer topology with LeWin blocks to balance local and global feature learning. Its local self-attention mechanism reduced complexity, yet the hierarchical structure often led to over-smoothing in fine texture regions. Overall, existing transformer-based SR methods reveal a trade-off between global context modeling and spatial detail preservation, a challenge that DRCT framework directly targets by introducing efficient dense residual connections across transformer layers for multi-image consistency.

\subsection{DRCT}

We build upon the Dense Residual Connection Transformer (DRCT) \cite{DRCT}, which was originally designed for single-image SR. DRCT mitigates the information bottleneck prevalent in deep transformer networks by introducing dense residual connections across layers, improving gradient flow and multi-level feature reuse. This enables effective fusion of local and global context without compromising spatial precision.

While DRCT has achieved state-of-the-art results in single-image SR, it has not been extended to the multi-image domain. Our proposed MI-DRCT addresses this gap by incorporating inter-image feature alignment and fusion, allowing the model to leverage complementary information across multiple low-resolution inputs. This design not only enhances fine-grained texture recovery but also improves robustness against noise and illumination variations typical of underground root imaging scenarios.

\subsection{Analysis of Root Images}

Root imaging has been traditionally dominated by classical image analysis approaches \cite{root_analysis1, root_analysis2, root_analysis3, root_analysis_divya}. Early methods relied on morphological operations or geometric fitting (e.g., cylinder or edge-based) to infer structural properties such as diameter, area, and length. While effective for isolated samples, these methods fail in complex root systems due to occlusion, noise, and poor contrast between roots and soil.

Ground-penetrating radar (GPR) techniques \cite{root_analysis_radar1, root_analysis_radar2, root_analysis_radar3, root_analysis_radar4, root_analysis_radar5, root_analysis_radar6} enable non-destructive subsurface imaging, but their performance is highly sensitive to soil texture, moisture, and root density, limiting generalizability. These dependencies make consistent quantitative analysis difficult across environmental conditions.

Recent advances in deep learning have reintroduced RGB-based underground imaging \cite{underoot1, underoot2}, leveraging enhancement models such as SR to recover structural details. However, most existing approaches remain preliminary \cite{root_analysis_deep1, root_analysis_deep2} and struggle with domain variability—particularly due to soil heterogeneity and moisture interference. Data acquisition is also slow and inconsistent, making robust model training challenging. Our MI-DRCT framework addresses these shortcomings by improving resolution and contrast through multi-image fusion, enabling more reliable feature extraction from adverse imaging conditions.

Li et al. \cite{sun2024matching} observed that lightweight SR models often suffer from severe accuracy degradation, while large-capacity models incur prohibitive computational costs. Their multi-scale feature selection network (MFSN) achieved a favorable trade-off, but still relied on single-image inputs, limiting its robustness under varying soil or illumination conditions. Similarly, Wu et al. \cite{wu2024multi} proposed a Multi-Scale Non-Local Attention Network (MSNLAN) to capture long-range pixel dependencies, but their architecture introduced redundancy across scales, increasing inference latency. These works collectively underscore the need for architectures that maintain spatial fidelity while remaining computationally efficient—principles central to our proposed MI-DRCT.

\section{Methodology}

Our method consists of four different components. These are Multi-Image RootCam for real plant root capture, Synthetic Root Image Generation for training, MI-DRCT for image enhancement and Root Image Analysis to obtain the root characteristics. We elaborate on each of these in the following subsections.

\subsection{Multi-Image RootCam}

\begin{figure}[t]
\centering
\includegraphics[width=3in]{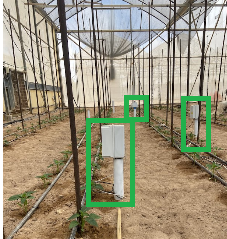}
\caption{RootCam deployed in an agriculture field. Note that there are three visible RootCam setups each highlighted with a green rectangle.}
\label{fig:cam}
\end{figure}

Multi-Image RootCam (CrystalVision, Samar, Israel), as shown in Figure \ref{fig:cam}, is an extension of RootCam \cite{root_analysis_divya}. It consists of a transparent tube, closed at the bottom, which is 100 cm in height with an external diameter of 60 mm and an internal diameter of 54 mm. The tube is installed vertically into the soil. The top part, which remains above the soil, is painted black on the inside to prevent light from entering and white on the outside to reflect excess light. The RootCam device is then carefully inserted into the tube for imaging (Figure \ref{fig:cam}).

The RootCam device itself includes a rotating sled that carries a high-resolution camera (2592 × 1944 pixels, 0.01 mm spatial resolution, 2500 dpi) and integrated LED light sources (white and UV). To avoid interfering with plant growth, illumination is only activated during image acquisition. The camera acquires images every 20 mm as it travels along the tube. Its positioning and movement (vertical and rotational) are controlled via the RootCam software interface, which also provides features like amount of image overlap, super-resolution, and real-time preview.

All captured images are stored on a built-in Raspberry Pi computer, which can be accessed either locally through a LAN or remotely via Wi-Fi, VNC, or Raspberry Pi Connect. The system also supports automatic uploading to cloud storage services such as Dropbox. Each image file contains metadata, including the tube depth, rotation step, and acquisition time. The device operates on a 12V DC power supply (5A), and the motion control enables sub-pixel positioning of the camera within the tube. This comprehensive setup allows for continuous, non-destructive monitoring of root dynamics at various depths with precise spatial and temporal resolution.

\begin{figure}[!t]
\centering
\includegraphics[width=3in]{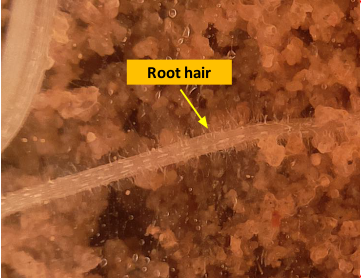}
\caption{A sample of root image showing root hairs captured by the RootCam.}
\label{fig:hair}
\end{figure}

\subsection{Synthetic Root Image Generation}
Although we have access to the RootCam setup, acquiring a sufficiently large number of real images for training a deep neural network, typically in the order of thousands, is challenging. The slow growth rate of roots and the need for diverse soil and lighting conditions limit large-scale real data collection. To overcome this constraint, we simulate underground conditions and generate synthetic root images. Specifically, we capture diverse real background samples using the RootCam system and employ statistical models to overlay roots of varying shapes and root hair densities. These synthetic images significantly expand the training set and improve model convergence.

\begin{figure}[!t]
\centering
\includegraphics[width=3in]{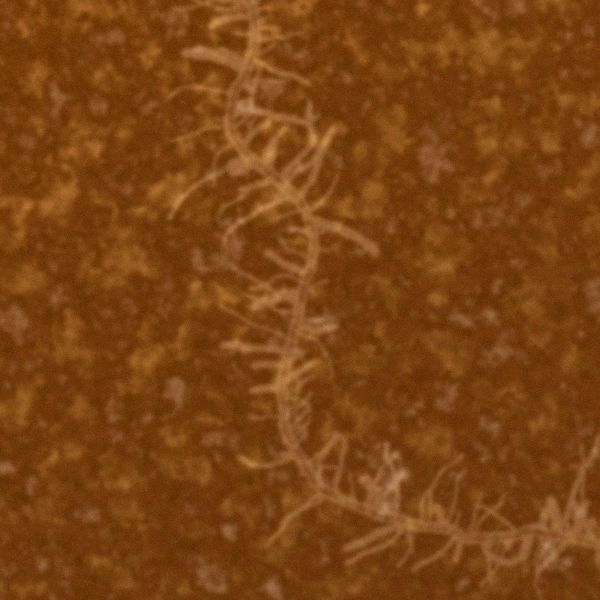}
\caption{An image generated by our synthetic root generation algorithm.}
\label{fig:synthetc_image}
\end{figure}

We first generate a very high resolution synthetic root image. Then multiple overlapping root images are generated by cropping such an image with controlled sub-pixel vertical shifts. This process involves the following steps:

\begin{itemize}
\item	Image Cropping: A sub-region of the synthetic root image, centered around the root structure, is cropped to focus on the area of interest.
\item		Vertical shifts are applied to the image by an odd number of pixels. Using an odd number of pixels for the shift ensures that when the image is subsequently downscaled by a factor of 2, a sub-pixel shift of 0.5 will be created. This introduces more unique variations in the shifted images, which is beneficial for the super-resolution task.
\item	Downscaling and Interpolation:
The shifted cropped images are downscaled using \verb|cv2.INTER_AREA| interpolation. It calculates the average pixel value over the area of each new pixel in the downscaled image. This helps to minimize aliasing artifacts and preserve image details during the downscaling, which is crucial for retaining the sub-pixel shift information.
\end{itemize}

Training on synthetically generated data not only compensates for the scarcity of real samples but also enhances the robustness and generalization capability of the MI-DRCT model. By simulating a wide range of lighting variations, soil textures, and root morphologies, the synthetic dataset exposes the model to diverse visual conditions that may not be adequately represented in the limited real dataset. This diversity enables the model to learn invariant features that are resilient to illumination changes, occlusions, and noise present in real RootCam images. Consequently, models pre-trained on synthetic data demonstrate improved stability and adaptability when fine-tuned or directly applied to real-world root imaging scenarios.

\subsection{MI-DRCT}

\begin{figure*}[!t]
\centering
\includegraphics[width=7in]{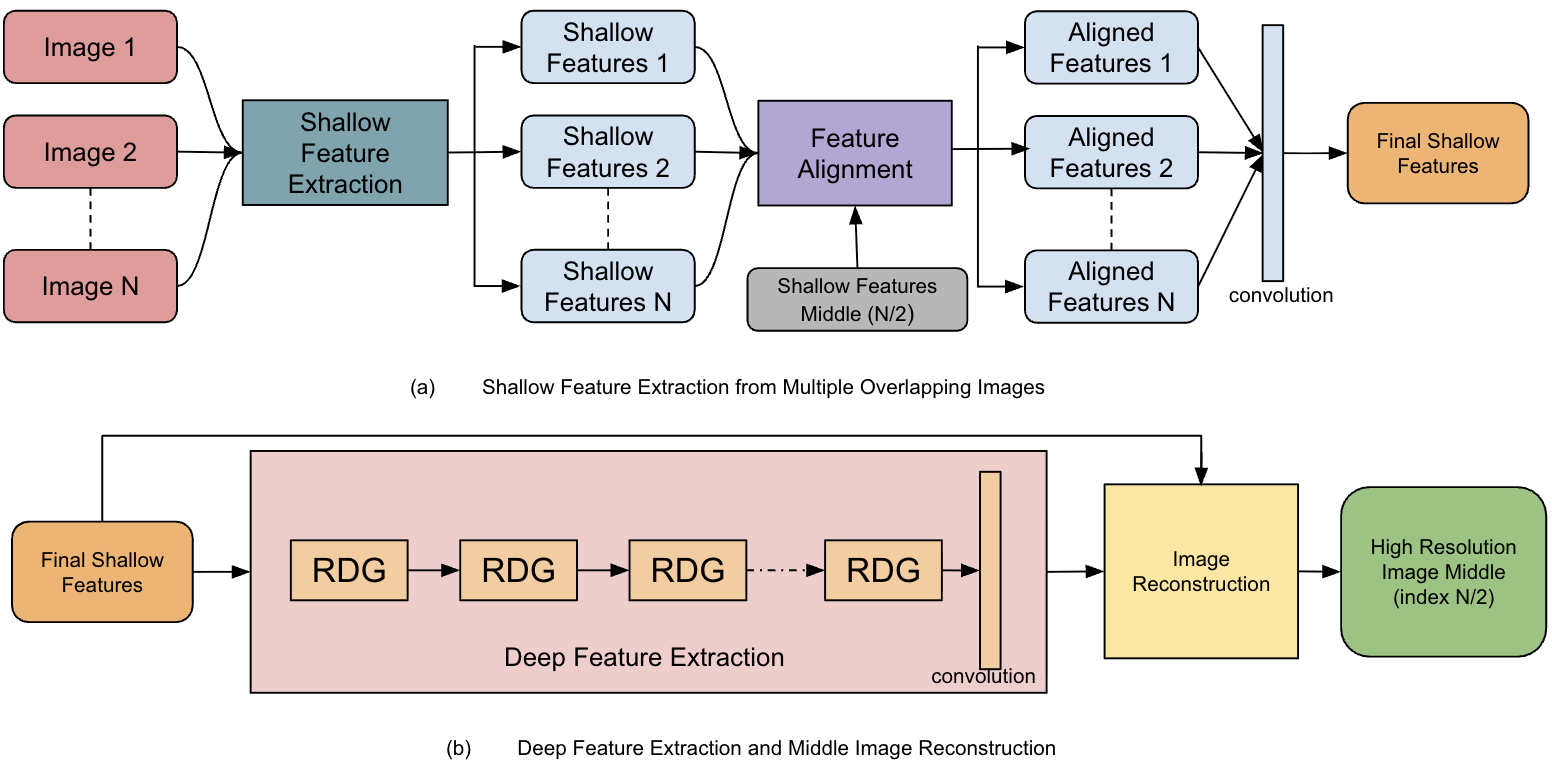}
\caption{Architecture of MI-DRCT. We have divided it into two stages. Step (a) does the shallow feature extraction and alignment of shifted features. Step (b) takes the combined shallow features as input and extracts the deep features using Residual Dense Group (RDG) blocks. The deep and shallow features are then combined to produce the super-resolution image.}
\label{midrct}
\end{figure*}

As we discussed in previous sections, the RootCam can be used to capture multiple overlapping images for plant roots and we also have a way to generate synthetic images that can simulate this setting. Now, we create an algorithm to enhance the image quality by exploiting the redundancy and sub-pixel information present in the overlapping images. We take DRCT \cite{DRCT} which is the current SOTA performing framework for single image super-resolution and modify it to utilize the redundancy and sub-pixel information in the underground root data.  The new framework is called Multi-Image Deep Residual Channel Transformation (MI-DRCT). The architecture of MI-DRCT is described in Figure \ref{midrct} .

 The  MI-DRCT framework comprises two primary stages, as illustrated in Figure \ref{midrct}. These stages are designed to exploit spatial redundancy across overlapping input images to enhance feature representation and ultimately improve super-resolution performance.

\subsubsection{ Shallow Feature Extraction and Alignment}
The first stage, depicted in Figure \ref{midrct}(a), involves the extraction and alignment of shallow features from multiple overlapping low-resolution (LR) input images. Each of the N input images is independently passed through a shared convolutional layer to extract initial shallow features. This shared layer ensures consistent feature representation across the image stack.

Once the initial features are obtained, the feature map corresponding to the central (or middle-index) image in the input sequence is designated as the reference feature map. The remaining N--1 feature maps are aligned to this reference using a Feature Alignment Module. This module is designed to handle small spatial misalignments across images caused by camera motion or parallax.

The alignment process leverages a Phase Correlation Module, which estimates the relative shift between the query and reference feature maps in the Fourier domain. Given the nature of the input (e.g., root images), we observe that the misalignment is predominantly along the vertical axis. Hence, only the vertical component of the estimated shift is considered for alignment. To enhance the precision of alignment, especially in the presence of sub-pixel shifts, both the query and reference feature maps are upsampled by a factor of 2 prior to phase correlation.

Following alignment, all feature maps are concatenated along the channel dimension. For N images, each with a feature dimension of f, the resulting concatenated tensor has a dimensionality of N×f. This concatenated feature map is then passed through a second convolutional layer to reduce the channel dimension back to f, yielding the Final Shallow Features.

\subsubsection{ Deep Feature Extraction and Image Reconstruction }
The second stage, shown in Figure \ref{midrct}(b), involves deeper feature extraction and image reconstruction. The Final Shallow Features are processed through a series of Residual Deep Feature Extraction Groups (RDGs), as originally proposed in the DRCT \cite{DRCT} framework. These RDGs are designed to extract complex hierarchical features that contribute to fine-grained detail restoration in the high-resolution (HR) output.

To preserve low-level information and promote gradient flow, a skip connection is employed that concatenates the shallow and deep features. This fused feature representation is subsequently fed into the Image Reconstruction Module, which generates the final super-resolved output corresponding to the central image in the input sequence.

It is important to note that all convolutional layers in the architecture utilize 3×3 kernels with an additional zero-padding. This configuration ensures that the spatial resolution (i.e., height and width) of the output tensor remains identical to that of the input at each convolutional stage. 

\subsection{Root Image Analysis}

To analyze the morphology of underground plant roots, we developed and implemented a deep learning-based instance segmentation model. This model was trained on a synthetic dataset specifically generated to simulate underground root structures, including both primary roots and fine root hairs. Given an input image of subterranean roots, the model produces pixel-wise instance segmentation masks that distinctly identify the root body and associated root hairs.

Following segmentation, the model computes the average area of root hairs in pixels. This pixel-based measurement is then converted to physical units (millimeters squared) using the known spatial resolution of the image. The resulting metric provides a quantitative estimate of root hair density, which can serve as a useful indicator for assessing overall plant health and nutrient uptake capability.

Figure \ref{fig:detect} presents the segmentation results on a sample output from MI-DRCT and other algorithms for a real root image input. As shown, the primary root structure is delineated in light blue, while the root hairs are distinctly marked in dark blue, enabling clear visual differentiation between these two anatomical components.

\section{Experiment Setup}
All experiments were conducted on a Windows-based workstation equipped with an NVIDIA RTX 3090 GPU. To train our proposed MI-DRCT model, we constructed a synthetic dataset consisting of 10,000 root image triplets for training and an additional 2,000 triplets for validation. Each triplet consisted of 3 images with the vertical shift accounting for $N=3$ in MI-DRCT (Figure \ref{midrct}). The number of overlapping images was set to 3 to maximize the number of independent samples for evaluation. This synthetic setup allows for a controlled experimentation and ensures consistency across evaluations. After these experiments on the synthetic data, we applied the model on real dataset consisting of 50 triplets of underground root images for a Bell Pepper plant. This experiment proves the effectiveness of the system in the real world setting.

The MI-DRCT model was trained for 15 epochs, and the model checkpoint yielding the best performance on the validation set was selected for final inference. During training, we employed a batch size of 8 and optimized the network using the Adam optimizer \cite{adam}. The initial learning rate was set to $1 \times 10^{-3}$, and no learning rate decay was applied.

Each input image was cropped to a spatial resolution of 64×64 pixels, and the model was trained to generate corresponding high-resolution outputs of size 128×128, corresponding to an upscaling factor of 2. Owing to its fully convolutional architecture, the MI-DRCT model is resolution-agnostic and can be seamlessly adapted to inputs of arbitrary dimensions during inference, without requiring architectural modifications.

For comparative evaluation, we employed the baseline DRCT model \cite{DRCT} using its default hyperparameter settings. Specifically, the embedding dimension was set to 96, the window size to 16, and the transformer configuration was maintained with depths and number of heads set to (6,6,6,6), as originally proposed. The upscaling factor was also fixed at 2 to maintain consistency with our proposed model’s output resolution.

\section{Results}

\begin{figure*}[!t]
\centering
\includegraphics[width=7in]{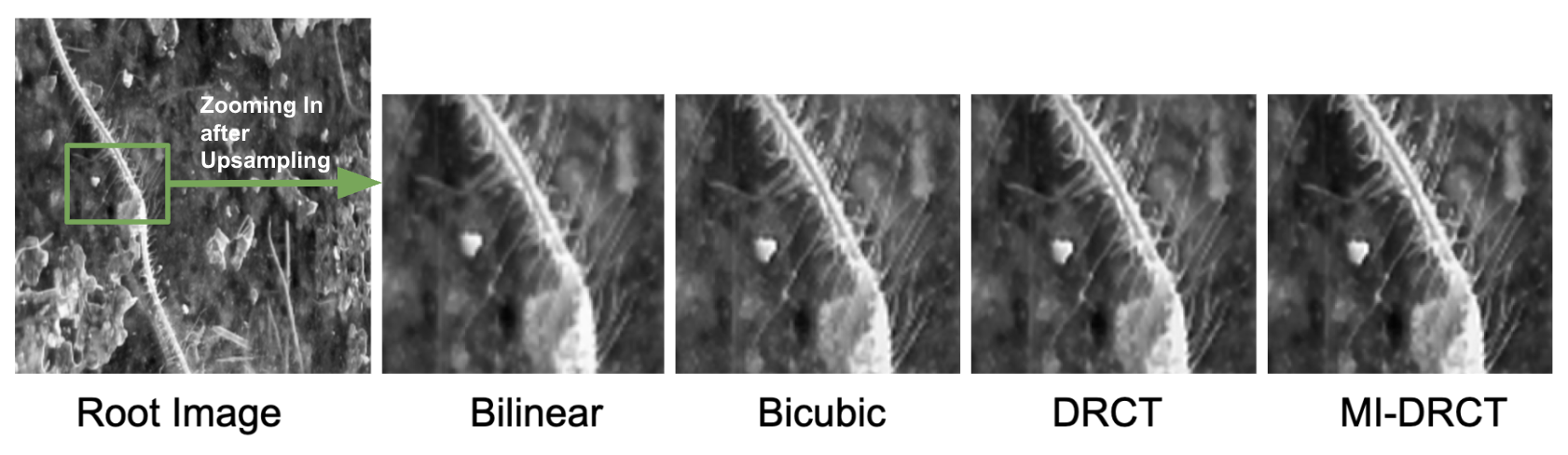}
\caption{Qualitative comparison of different super-resolution techniques on a real image captured by RootCam for a Bell Pepper plant. We can see the higher contrast in the box for MI-DRCT which enhances the root hairs. DRCT also has good contrast but slightly lower quality than the MI-DRCT.}
\label{image_comp}
\end{figure*}

As mentioned in the previous sections, we train our model on synthetic data. We test the multi image super resolution model on the withheld synthetic data and compare it with other state-of-the-art single and multi-image super-resolution algorithms. We also test and compare these models on the real root images captured using RootCam. We then detect and analyze the root characteristics on the enhanced images. The CLIP-IQA metric in our experiments compares an image’s embedding with two fixed textual prompts, ``Good photo” and ``Bad photo”. It computes cosine similarities between the image and both prompts and outputs the probability that the image is closer to the positive prompt, representing its perceptual quality score. 

\subsection{Results on Synthetic Root Dataset}
We compared our new MI-DRCT algorithm with the existing approaches on single image and multi-image super resolution on both with-reference (MSE, PSNR and SSIM) and no-reference (BRISQUE and CLIP-IQA) image quality metrics.

\begin{table}[H]
\caption{Comparison of the proposed MI-DRCT framework with other approaches on reference and no-reference image quality metrics (synthetic test dataset).}
\label{tab:table1}
\centering
\setlength{\tabcolsep}{4pt} 
\renewcommand{\arraystretch}{1.0} 
\begin{tabular}{c|ccccc}
\textbf{Framework} & \textbf{MSE $\downarrow$} & \textbf{PSNR} & \textbf{SSIM} & \textbf{BRISQUE $\downarrow$} & \textbf{CLIP-IQA} \\
\hline
Bilinear & 25.28 & 36.80 & 0.93 & 46.92 & 0.32\\
Bicubic & 19.84 & 38.05 & 0.94 & 43.79 & 0.36\\
DRCT & 18.14 & 38.22 & \textbf{0.95} & 44.23 & 0.38\\
\textbf{MI-DRCT} & \textbf{16.31} & \textbf{38.96} & \textbf{0.95} & \textbf{43.13} & \textbf{0.39}\\
\end{tabular}
\end{table}

In Table \ref{tab:table1}, we present the comparison of super resolution quality from our framework (MI-DRCT) with the existing approaches. It can be clearly seen that the MI-DRCT framework outperforms all the existing approaches on both no-reference and with reference image quality metrics. We can see that the MI-DRCT has lower mean squared error (MSE), and higher peak signal to noise ratio (PSNR) and structural similarity index measure (SSIM) beating the current SOTA approach of DRCT \cite{DRCT} and other baselines on with reference quality matrices. Similar behaviour is observed for no-reference image quality matrices of BRISQUE \cite{BRISQUE} and CLIP-IQA \cite{clipiqa}, where MI-DRCT achieves the best results.

Note that the DRCT approach is the current SOTA benchmark for single image super-resolution and using multiple images with sub-pixel shifts to enhance the information via MI-DRCT shows a promising direction to extrapolate other single image super-resolution framework to work for multi-image settings.

\subsection{Results on Real Root Dataset}

\begin{figure*}
\centering
\includegraphics[width=7.1in]{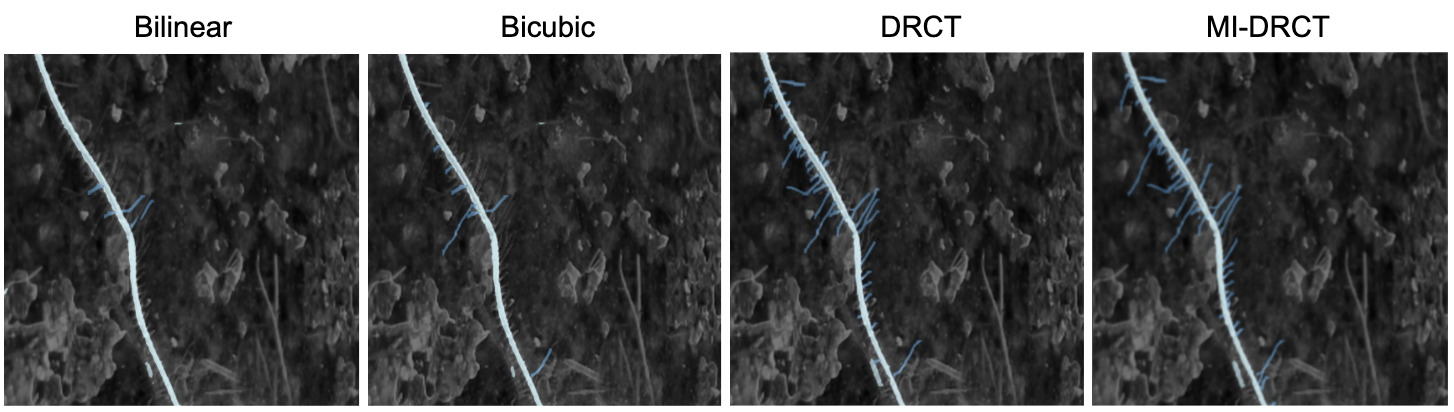}
\caption{Comparison of image segmentation algorithm performance with different super-resolution approaches. The MI-DRCT enhanced image enables more accurate automatic detection of both root and root hairs.}
\label{fig:detect}
\end{figure*}

We also compared the results from MI-DRCT framework trained on synthetic data with other methods on the real underground root test images obtained via the Multi-Image RootCam. The real images are from a Bell Pepper plant. Here, we only have no-reference image quality metrics since there is no reference image for super-resolution evaluation. This limits our comparison to BRISQUE and CLIP-IQA scores. The results are presented in Table \ref{tab:table3}.

\begin{table}[H]
\caption{Table showing comparison of the proposed MI-DRCT super resolution framework with other approaches on no-reference image quality metrics on the real image test dataset.\label{tab:table3}}
\centering
\begin{tabular}{c|cc}
\textbf{Framework} & \textbf{BRISQUE $\downarrow$} & \textbf{CLIP-IQA } \\
\hline
Bilinear & 46.48 & 0.30\\
Bicubic & 44.61 & 0.31\\
SwinIR \cite{liang2021swinir} & 46.03 & 0.33\\
BSRGAN \cite{bsrgan} &46.57 & 0.36\\
DRCT \cite{DRCT} & 45.33 & \textbf{0.38}\\
\textbf{MI-DRCT} & \textbf{44.50} & \textbf{0.38}\\
\end{tabular}
\end{table}

We can clearly see that the MI-DRCT algorithm has the best performance on the real data as well. It has the lowest BRISQUE value of 44.50 and CLIP-IQA score of 0.38. The original DRCT algorithm has the same CLIP-IQA score but has higher BRISQUE. Figure \ref{image_comp} presents a qualitative comparison of original and enhanced images from various super-resolution algorithms. MI-DRCT visibly improves root hair visualization over bilinear and bicubic methods. While DRCT and MI-DRCT appear visually similar, quantitative metrics confirm MI-DRCT's superior performance, highlighting the benefit of using multiple shifted images.

It is important to note that the performance of synthetic and real test dataset closely align with similar BRISQUE and CLIP-IQA scores. As getting the real data is hard, this result is important in showing that the synthetically generated root data is also effective for training the automatic root hair detection models.

\subsection{Detection and Analysis of Plant Root}

\begin{table*}
\centering
\caption{Table showing the characteristics of the identified root hairs in Figure \ref{fig:detect}.}
\label{tab:table4}
\begin{tabular}{l|cccc}
\textbf{Root Trait} & \textbf{Bilinear} & \textbf{Bicubic} & \textbf{DRCT} & \textbf{MI-DRCT}\\
\hline
Root Count (\textbf{Human Expert: 1}) & 1 & 1 & 1 & 1 \\
Root Hair Count (\textbf{Human Expert: 77}) & 4 & 8 & 38 & \textbf{44} \\
Total Root Hair Length ($mm$) & 7.88 & 14.71 & 58.54 & 71.47 \\
Average Root Hair Length ($mm$) & 1.97 & 1.83 & 1.54 & 1.62 \\
Average Root Hair Area ($mm^{2}$) & 0.29 & 0.25 & 0.20 & 0.19 \\
\end{tabular}
\end{table*}

The proposed MI-DRCT algorithm significantly enhances the segmentation model’s ability to accurately detect fine-scale structures, particularly root hairs, which are often difficult to distinguish due to their small size and low contrast in raw images. As illustrated in Figure \ref{fig:detect}, the segmentation algorithm detects only a few root hairs with bilinear and bicubic interpolation. With DRCT, the root hair detection improves drastically and we detect 38 root hairs. In contrast, with MI-DRCT we get 44 root hairs out of the 77 root hairs marked by human expert.
Since the spatial calibration of the imaging system (in mm) is already known, the segmentation output can be reliably mapped to physical units. This allows for the automated extraction of relevant metrics like root hair length and area. The detailed analysis is presented in Table \ref{tab:table4}. With MI-DRCT, we get a total root hair length of 71.47 mm and an average root hair length of 1.62 mm. We also perform similar analysis for other super-resolution algorithms as well. Root hair length and area are too fine and even expert analysis leads to heavy inaccuracies. Our frameworks capability to capture fine morphological features are essential for studying plant physiology, nutrient uptake, and overall root system architecture.

\section{Ablation Studies}

\subsection{Importance of sub-pixel alignment}

We conduct an ablation study to determine the importance of sub-pixel shifting (upscaling by 2 before shallow feature alignment) module in the MI-DRCT approach. The results are presented in Table \ref{tab:table2}.

\begin{table}[H]
\caption{Table showing the importance of accounting for sub-pixel shifts in the  MI-DRCT super resolution framework on synthetic underground root image dataset.\label{tab:table2}}
\centering
\begin{tabular}{p{1.55cm}|p{0.85cm}p{0.52cm}p{0.52cm}p{1.49cm}p{1.30cm}}
\textbf{Framework} & \textbf{MSE $\downarrow$} & \textbf{PSNR} & \textbf{SSIM} & \textbf{BRISQUE $\downarrow$} & \textbf{CLIP-IQA} \\
\hline
\textbf{MI-DRCT (no subpixel)} & 17.65 & 38.45 & 0.95 & 43.80 & 0.37\\
\textbf{MI-DRCT} & 16.31 & 38.96 & 0.95 & 43.13 & 0.39\\
\end{tabular}
\end{table}

Table \ref{tab:table2} highlights the critical role of accounting for sub-pixel shifts within the proposed MI-DRCT framework, evaluated on a synthetic underground root image dataset. Two variants of the MI-DRCT framework are compared, one that directly aggregates multi-image inputs without considering sub-pixel displacements, and another that incorporates sub-pixel alignment across frames prior to fusion.

The results show that incorporating sub-pixel shifts leads to a notable improvement in the overall image reconstruction quality. Specifically, there is a 7.5\% reduction in Mean Squared Error (MSE), indicating lower pixel-wise reconstruction error. The SSIM (0.95) and PSNR (~38.9 dB) remain consistent across both variants, suggesting comparable structural and signal information. BRISQUE (a measure of naturalness) decreases from 43.80 to 43.13 (lower is better), and the CLIP-IQA score increases from 0.37 to 0.39, demonstrating better alignment with semantically meaningful image content. Overall, the sub-pixel aware model shows superior performance. 

These improvements confirm that sub-pixel shifts across multiple low-resolution frames capture non-redundant and complementary spatial details that can be leveraged to reconstruct higher-quality super-resolved images. Without compensating for these shifts, the aggregated frames provide largely overlapping information, limiting the potential gain from multi-image super-resolution. Thus, sub-pixel alignment is essential for effectively utilizing the additional inter-frame information present in multi-image inputs.

\section{Conclusion}
This paper presents a novel framework designed to enhance the detection and analysis of plant roots and root hairs by leveraging multiple overlapping RGB images. Root and root hair phenotyping play a critical role in plant biology and agricultural research, yet capturing detailed and accurate imagery of underground root systems remains a significant challenge due to occlusion, low contrast, and soil interference. To address this, we propose a multi-image enhancement approach that integrates information from several RGB views to reconstruct a high-quality representation of the subterranean root structure. Our method enables more accurate localization and counting of root hairs without the need for invasive imaging techniques.

At the core of our framework is the MI-DRCT (Multi 
Image Deep Residual
Channel Transformation) algorithm, a deep learning-based super-resolution model trained on synthetically generated datasets that simulate complex underground conditions. Training with synthetically generated root images enhances the robustness and generalization of the MI-DRCT model by exposing it to a wider variety of textures, lighting conditions, and root structures. This diversity enables the model to learn more invariant features, improving its performance and stability on real RootCam data. Through extensive experimentation, we demonstrate that MI-DRCT outperforms existing state-of-the-art algorithms in reconstructing fine-scale root structures. The model exhibits superior performance in both synthetic environments and real-world imaging conditions, producing high-fidelity visual outputs that facilitate downstream tasks such as root segmentation, feature extraction, and root hair quantification.

Our results demonstrate that integrating multiple low-resolution RGB images in a structured and learnable framework can lead to substantial improvements in both the visual fidelity and analytical utility of root imagery. By leveraging the complementary information present across multiple views, the proposed MI-DRCT framework effectively reconstructs high-resolution root structures with enhanced clarity and detail. This is particularly beneficial for non-invasive root imaging, where preserving fine structural features is critical for downstream analysis. The MI-DRCT framework also facilitates more accurate and reliable interpretation of root architectures, which is essential for quantitative phenotyping tasks such as root length estimation, branching pattern detection, and growth monitoring. Furthermore, its modular design and scalability make it well-suited for deployment in high-throughput plant phenotyping pipelines, where large volumes of root imagery must be processed efficiently and consistently.

Overall, the proposed method represents a significant advancement in root imaging technologies, offering a practical and robust solution for researchers and agronomists aiming to understand root system dynamics under various soil and environmental conditions.

\section*{Acknowledgments}
The project was supported by the Ministry of Agriculture and Food Security: Development of an automatic system for characterizing the root system for optimal irrigation and fertilization. This research was also partially funded by the Planning and Budgeting Committee of the Council for Higher Education in Israel (VATAT), through the Israeli Center for Digital Agriculture program.

\bibliographystyle{IEEEtran}
\bibliography{references}

\end{document}